\documentclass{article} 
\usepackage{iclr2022_conference,times}

\usepackage{amsmath,amsfonts,bm}

\def\eqref#1{equation~\ref{#1}}

\def\1{\bm{1}}

\DeclareMathAlphabet{\mathsfit}{\encodingdefault}{\sfdefault}{m}{sl}
\SetMathAlphabet{\mathsfit}{bold}{\encodingdefault}{\sfdefault}{bx}{n}

\newcommand{\E}{\mathbb{E}}

\DeclareMathOperator*{\argmax}{arg\,max}

\usepackage{hyperref}
\usepackage{url}
\usepackage{algorithm}
\usepackage{tabularx}
\usepackage{graphicx}
\usepackage{algorithmic}
\usepackage{subcaption}
\usepackage{amsmath,amsthm,amssymb}
\usepackage[ruled,vlined,algo2e]{algorithm2e}
\usepackage{pdflscape}

\usepackage{authblk}

\title{Target Entropy Annealing for Discrete Soft Actor–Critic}
\iclrfinalcopy

\author[1,\thanks{Correspondence to: yaoshenx@uci.edu}]{Yaosheng Xu}
\author[1]{\ Dailin Hu}
\author[1]{\ Litian Liang}
\author[1]{\ Stephen McAleer}
\author[2]{\ Pieter Abbeel}
\author[1]{\ Roy Fox}
\affil[1]{Department of Computer Science, University of California, Irvine}
\affil[2]{Department of Electrical Engineering and Computer Science, \linebreak University of California, Berkeley}

\begin{document}

\maketitle

\begin{abstract}
    Soft Actor-Critic (SAC) is considered the state-of-the-art algorithm in continuous action space settings. It uses the maximum entropy framework for efficiency and stability, and applies a heuristic temperature Lagrange term to tune the temperature $\alpha$, which determines how "soft" the policy should be. It is counter-intuitive that empirical evidence shows SAC does not perform well in discrete domains. In this paper we investigate the possible explanations for this phenomenon and propose Target Entropy Scheduled SAC (TES-SAC), an annealing method for the target entropy parameter applied on SAC. Target entropy is a constant in the temperature Lagrange term and represents the target policy entropy in discrete SAC. We compare our method on Atari 2600 games with different constant target entropy SAC, and analyze on how our scheduling affects SAC.
\end{abstract}

\section{Introduction}
    Deep reinforcement learning (RL) algorithms are capable of learning good behavior policies in a wide range of environments \citep{silver2016Mastering, DBLP:journals/corr/MnihKSGAWR13, schulman2015TRPO, gu2017robotic}. In environments with continuous action spaces, Soft Actor–Critic (SAC)~\citep{haarnoja2018soft} has been shown to learn robustly for control tasks in simulated environments and real-world robots \citep{haarnoja2018robotic}. Motivated by the Maximum-Entropy (MaxEnt) RL theory~\citep{ziebart2010Model, todorov2008optimalControl, olsson2005maxent, toussaint2009robot, Fox2016Taming, levine2018reinforcement}, SAC simultaneously trains a critic to evaluate the free-energy (value–entropy tradeoff) of an actor policy, and the actor to imitate the softmax policy induced by the critic. To set the value–entropy tradeoff coefficient, the \emph{temperature} $\alpha$, SAC optimizes an objective in which $\alpha$ is the Lagrange multiplier for the constraint that the average actor's entropy $\mathcal{H}$ is no less than a given target entropy $\bar{\mathcal{H}}$.
    
    
    SAC in discrete action spaces has not shown the same success as in continuous action spaces~(\citealt{christodoulou2019soft}; and see its implementation in the RLlib Python package, \citealt{DBLP:journals/corr/abs-1712-09381}). \cite{christodoulou2019soft} suggests setting the target entropy $\bar{\mathcal{H}}$ to 98\% of the maximum possible policy entropy. While this configuration can lead to stable learning of good policies in some environments, in most scenarios it greatly underperforms. In most environments, there simply is no good policy that satisfies this target entropy constraint, and reaching a high temperature $\alpha$ that induces a high-entropy actor is detrimental to achieving high policy value, since it would be close to random. On the other hand, setting a low constant target entropy restraint forces SAC to quickly decrease $\alpha$ at early stage, and consequently with $\alpha \to 0$ the MaxEnt learning objective returns to the classic RL learning objective. The actor would attempt to learn the logits of a deterministic greedy policy, and the critic will try to evaluate it. Because the logits of a deterministic policy saturate, this algorithm is prone to early overfitting that is difficult to “unlearn”. Note that this issue is less severe in continuous action spaces, where the greedy action (the mean of the Gaussian policy distribution) is largely decoupled from the policy entropy (completely determined by the policy variance). 
        

    Selecting the correct value for the target entropy is evidently essential for obtaining good results in SAC. Using a constant target entropy throughout training requires potentially heavy fine-tuning for best performance. Alternatively, we look into automated tuning methods for the target entropy value. In this paper, we propose the Target Entropy Scheduled SAC (TES-SAC), a heuristic scheduling method applied on SAC to reach appropriate temperature values during training by gradually dropping the target entropy $\bar{\mathcal{H}}$, using the policy entropy as a signal. We initialize the target entropy to be the maximum possible entropy, $\log |A|$, and decrease it by a constant factor as soon as the average policy entropy stabilizes around the target. In this way, the average policy entropy will be high when training starts, and decreases gradually as training continues, reducing early overfitting to the insufficiently trained critic.

    While a similar approach can be applied to Soft Q-Learning (SQL)~\citep{Fox2016Taming, DBLP:journals/corr/HaarnojaTAL17}, a MaxEnt RL algorithm in which the policy is directly computed from a value network, we find that an actor policy network can stabilize training. Intuitively, a sudden change in the temperature $\alpha$ in SQL immediately causes a shift in the softmax policy, which creates a fast moving target for the value network training process. In contrast, the policy network in SAC takes time to adapt to a changing temperature, creating a more stable target for the value network.
    
    We analyze the problems with using a constant target entropy in SAC and propose TES-SAC in Section \ref{schedulingForTE}. We experiment on the Atari 2600 benchmark to compare our scheduled target entropy method with constant target entropy discrete SAC in Section \ref{exp} ~\citep{DBLP:journals/corr/MnihKSGAWR13}.
    Note that we are not trying to fine-tune TES-SAC to present state-of-the-art performance, but rather to provide empirical evidence that TES-SAC solves some of the issues with SAC using a constant target entropy in discrete settings. We also discuss the possibility of applying our target entropy scheduling method to SQL in comparison with SAC in Section \ref{discussions}.
    
\section{Preliminaries}
\label{preliminaries}
    \subsection{Notation}
    We mainly focus on discrete action spaces, with a Markov decision process (MDP) defined by $(\mathcal{S}, \mathcal{A}, p, r)$, where $\mathcal{S}$ is the state space, $\mathcal{A}$ is the discrete action space, $p(s'|s,a)$ is the state transition probability for current state $s \in \mathcal{S}$, action $a \in \mathcal{A}$, and next state $s' \in \mathcal{S}$, and $r(s, a)$ is the reward given an action-state pair. We want to learn a stochastic policy $\pi(a|s)$ that outputs action probability given a state. An optimal policy $\pi^*$ should maximize the expected discounted return $R = \sum_{t\geq0}\gamma^t r(s_t, a_t)$, where $\gamma$ is a discount factor with range [0, 1]. 
    \subsection{Soft Actor-Critic}
    Different from the above standard RL objective, Soft Actor-Critic \citep{haarnoja2018soft} uses an entropy-regularized objective \citep{ziebart2010Model}
    \begin{equation}
        \pi^* = \argmax_{\pi}\sum_{t\ge0} \gamma^t \E_{(s_t, a_t)\sim p_{\pi}}\big[r(s_t, a_t) + \alpha \mathbb{H}[\pi(\cdot |s_t)]\big] \label{eq:softrl}
    \end{equation}
    where $p_\pi(s_t, a_t)$ is the distribution over the state and action at time $t$ induced by rolling out the policy $\pi$ in the MDP, $\mathbb{H}[\pi(\cdot |s_t)]$ is the policy entropy at state $s_t$, and the \emph{temperature} $\alpha$ controls the trade-off between the entropy term and the expected rewards. When $\alpha \to 0$, the maximum entropy objective becomes the standard RL objective.
    
    To maximize this objective, SAC iterates between (1) updating a critic value function $Q_\theta(s, a)$ to match a soft Bellman backup target, and (2) minimizing the Kullback–Leibler (KL) divergence between an actor $\pi_\phi$ and the soft-greedy policy. 
    
   \textbf{Soft Bellman backup.} SAC updates the soft Q-function, parametrized by $\theta$, by minimizing the soft Bellman error for a state $s$, an action $a$ taken in $s$, the obtained reward $r$, and the next state $s'$
    \begin{equation}
        J_Q(\theta) = \tfrac{1}{2}(r + \gamma V_{\bar{\theta}}(s') - Q_{\theta}(s, a))^2,
        \label{qloss}
    \end{equation}
    where the next-step target value
    \begin{equation}
        \label{VFunc}
        V_{\bar{\theta}}(s') = \E_{(a' | s') \sim \pi_{\phi}} [Q_{\bar{\theta}}(s', a') - \alpha \log \pi_{\phi}(a' | s')]
    \end{equation}
    is computed from a target network $Q_{\bar{\theta}}$ copied periodically from the critic $Q_\theta$. The experience $(s, a, r, s')$ is obtained by sampling a replay buffer replenished by rollouts of the actor $\pi_\phi$. 

    \textbf{Policy update.}
    The policy network, parametrized by $\phi$ is a distillation of the softmax policy induced by the Q-function, and is updated by minimizing the KL-divergence
    \begin{equation}
        \mathrm{D}_{KL}\left[ \pi_\phi (\cdot | s) \left\| \frac{\exp\left(\frac{1}{\alpha}Q_\theta(s, \cdot)\right)}{Z_\theta(s) } \right. \right]
    \end{equation}
    over the parametric family of policies $\pi_\phi$, which in the continuous case is a Gaussian action distributions with mean and log-variance generated by a neural network. The partition function $Z_\theta(s)$ is a normalizer that can be ignored because it is action-independent. The resulting actor loss for a state $s$ sampled from the replay buffer is
    \begin{equation}
        J_{\pi}(\phi) = \E_{(a | s) \sim \pi_{\phi}}[\alpha \log (\pi_{\phi}(a_t | s_t)) - Q_{\theta}(s_t, a_t)].
        \label{ploss}
    \end{equation}
    
    SAC uses the temperature Lagrange term to tune the temperature $\alpha$ \citep{haarnoja2018soft_app}. Equation (\ref{eq:softrl}) can be viewed as the Lagrangian of the objective to find a policy with maximum expected return that at the same time satisfies an entropy constraint
    \begin{align*}
        \max_{\pi}\E_{p_\pi}\left[\sum_{t\ge0} \gamma^t r(s_t, a_t)\right]
    \end{align*}
    \begin{equation}
        s.t.  \;\;\; \sum_{t \ge 0} \gamma^t \E_{(s_t,a_t) \sim p_{\pi}}[-\log(\pi(a_t|s_t))] \ge \bar{\mathcal{H}},
    \end{equation}
    where $\bar{\mathcal{H}}$ is an externally selected threshold expected entropy. While the standard formulation (\ref{eq:softrl}) omits the constant $\bar{\mathcal{H}}$, optimizing for the temperature $\alpha$ involves minimizing
    \begin{equation}
        \label{alphaloss}
        J(\alpha) = \E_{(a | s) \sim \pi_t}[\alpha(-\log\pi_t(a_t|s_t) - \bar{\mathcal{H}})],
    \end{equation}
    where $\bar{\mathcal{H}}$ is thus called the \emph{target entropy}. The intuition is for the temperature to dynamically increase or decrease to encourage the policy entropy to approach the target entropy. 

    SAC can be straightforwardly applied to discrete action spaces~\citep{christodoulou2019soft}. One improvement that can be obtained in the discrete case is to directly calculate policy-expected values in (\ref{VFunc}), (\ref{ploss}), and (\ref{alphaloss}) as $\E_{a \sim \pi} f(s, a) = \sum_a \pi(a|s)f(s, a)$.
    This change reduces the variance by calculating the true expectation instead of sampling from the policy.
    In addition, while the policy action distribution in continuous Soft Actor-Critic is limited to a specific parametric family, usually Gaussian distributions, in discrete SAC we can represent any categorical action distribution by generating its logits. 
    

\section{Target Entropy Scheduled Soft Actor Critic}
\label{schedulingForTE}

    In this section, we first analyze the drawbacks of constant target entropy, then we present Target Entropy Scheduled SAC (TES-SAC) and explain in detail how we use this method to tune the policy entropy. Intuitively, in early stages of training, the policy is relatively random; as training proceeds it becomes increasingly deterministic. We should therefore drop the target entropy in a proper way that is neither too fast nor too slow: if we drop it too fast, the policy will be deterministic when the training is still in an early stage, and if too slow, the policy will remain stochastic for an undesirably long time. 
    
    \subsection{SAC with Constant Target Entropy}
    \label{constTE}
    In the original SAC temperature Lagrange term, target entropy is set to be a constant. In this section, we describe the issues with using a constant target entropy, and why the choice of the constant is extremely environment dependent.
    
    The temperature Lagrange term suggests that the policy entropy will try to approach the target entropy. Therefore when the target entropy is set to be a large constant, correspondingly the policy entropy will also be high. A policy with high policy entropy is similar to a random policy, which is undesirable in most cases. If we set the target entropy too small, the policy entropy will quickly drop, trying to hit the target, and become overly deterministic in an early stage of training. Moreover, if the target entropy is setted lower than the minimum achievable entropy level (see Section \ref{analysis}), $\alpha$ will exponentially decrease to zero. The soft Bellman loss in \eqref{qloss} and policy loss in \eqref{ploss} show that when $\alpha$ is small, the entropy term disappears, indicating the actor would learn the logits of a deterministic greedy policy, and the critic would evaluate it. The algorithm would easily overfit in an early stage and would be hard to “unlearn”, because the logits of a deterministic policy would saturate.
    As we explain in more detail in Section \ref{section:SAC_exp}, we still use a policy network in spite of this problem because it stabilizes the policy when dropping the target entropy. When dropping the target entropy, $\alpha$ will also drop abruptly, which will cause an immediate shift in the softmax policy, but a policy network will still need time to learn that shift which gives the value network a more stable target.

    Why not choose a constant target entropy that is neither too high nor too low? It is possible that certain constant target entropies perform well in some environments. Such a constant, however, will be extremely dependant on the dynamics of the environment and sensitive to noise. Tuning the best constant target entropy for each environment is expensive in computation and data. Our TES-SAC tries to solve the above problems.
    
    
    \subsection{Exponential Moving Window Scheduling}
    \label{expSchedule}
    Our TES-SAC scheduling method checks if the policy entropy has become stable and if it has approached the target entropy. If the policy entropy satisfies these requirements, we shrink the current target entropy by a factor. When calculating the policy entropy for the current iteration, we have
    
    \begin{equation}
        \label{entropy}
        e_i = -\E_b\Bigg[\sum_a{\pi(a|s)\log \pi(a|s)}\Bigg],
    \end{equation}
    
    where $e_i$ stands for the policy entropy for iteration $i$, $\pi(a|s)$ is the policy action distribution, and $E_b$ is the expectation over the mini-batch.
    
    We record the exponential moving mean $\hat{\mu}_i$ and the exponential moving standard deviation $\hat{\sigma}_i$ of the policy entropy, and calculate the exponential moving window of the policy entropy \citep{finch2009Incremental} such that
    \begin{equation}
        \hat{\mu}_i = \lambda \cdot \hat{\mu}_{i-1} + (1-\lambda) \cdot e_i,
    \end{equation}
    \begin{equation}
        \hat{\sigma}_i =\sqrt{ \lambda \cdot (\hat{\sigma}^2_{i-1} + (1 - \lambda) \cdot (e_i - \hat{\mu}_{i-1})^2)}.
    \end{equation}
        
    
    Here $i$ indicates the sequence in the exponential moving window. If $\hat{\mu}_i$ is close to the target entropy within the mean threshold, and $\hat{\sigma}_i$ is smaller than the standard deviation threshold, we multiply the target entropy by a constant factor. We describe this scheduling process in Algorithm 1, Appendix \ref{scheduling}.

\section{Experiments}
    \label{exp}
    We show through empirical experiments that TES-SAC can learn faster for some environments and provide domain-generalized hyperparameters compared to constant target entropy. For each environment, we compare our TES-SAC method with SAC with constant target entropy $\bar{\mathcal{H}} = C \cdot log|A|$, where $C \in \{0.98, 0.5, 0.01\}$. We experiment on several classical control tasks, as well as the Atari 2600 games. 
    
    \subsection{Overall Performance}

\begin{figure}[H]
            \centering
            \includegraphics[width=0.9\textwidth]{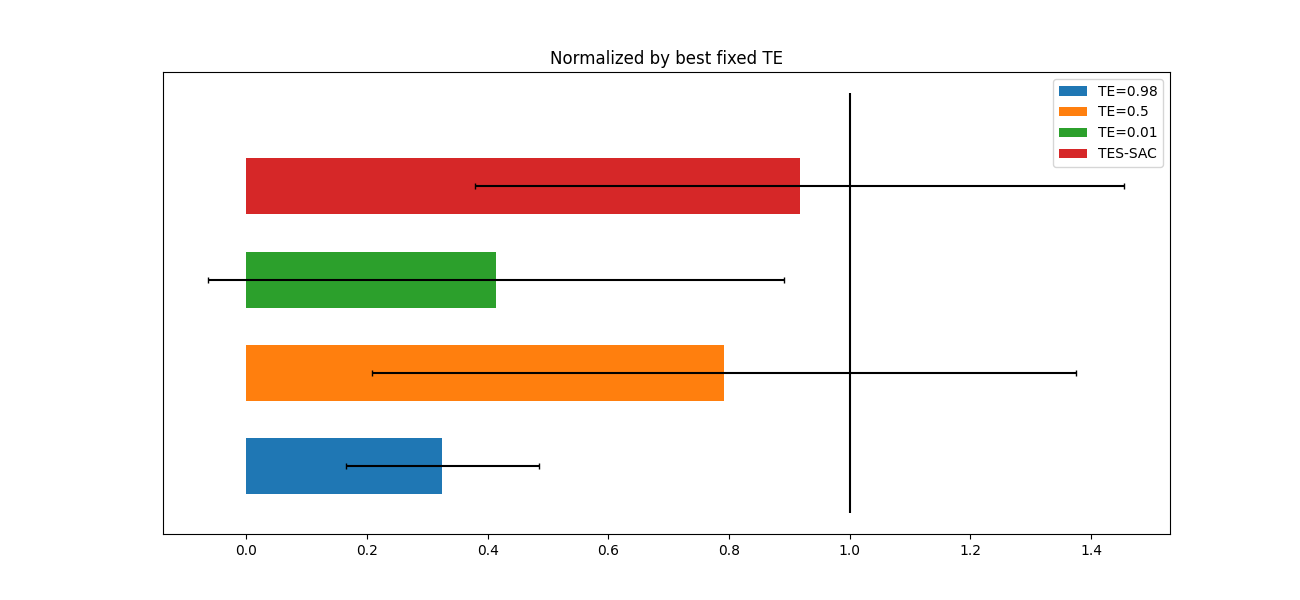}
            \caption{Performance of TES-SAC, where the range of worst-to-best fixed constant target entropy SAC is normalized to [0, 1], averaged over 24 environments. Error bars are plotted over three runs.}
            \label{fig:averaged_normalized}
        \end{figure}

\begin{figure}[H]
            \centering
            \includegraphics[width=0.24\textwidth]{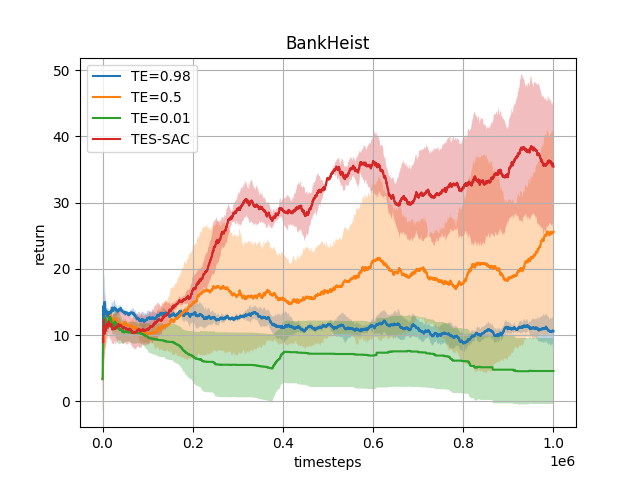}
            \includegraphics[width=0.25\textwidth]{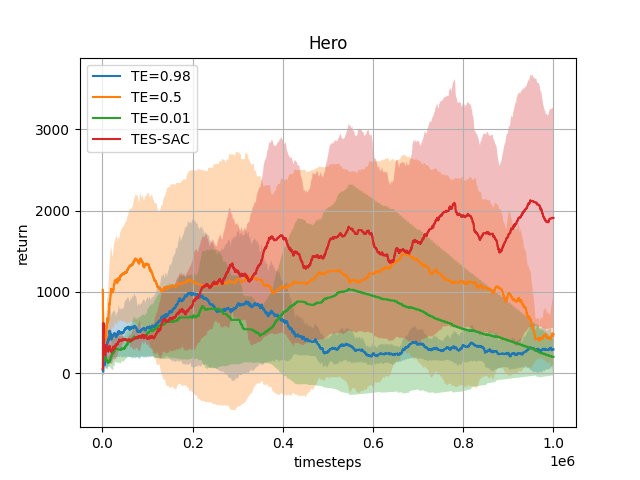}
            \includegraphics[width=0.24\textwidth]{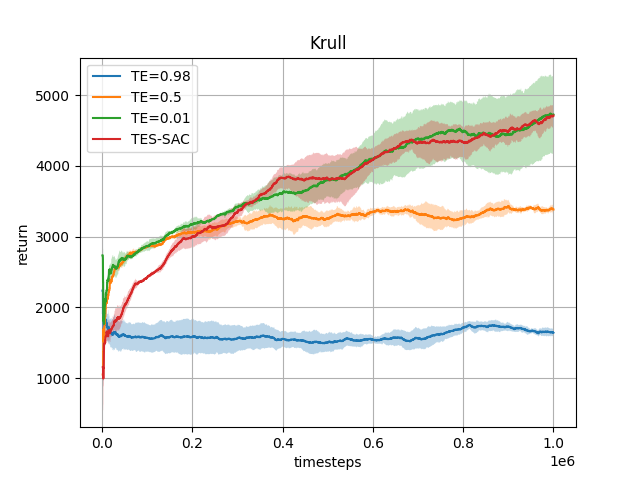}
            \includegraphics[width=0.25\textwidth]{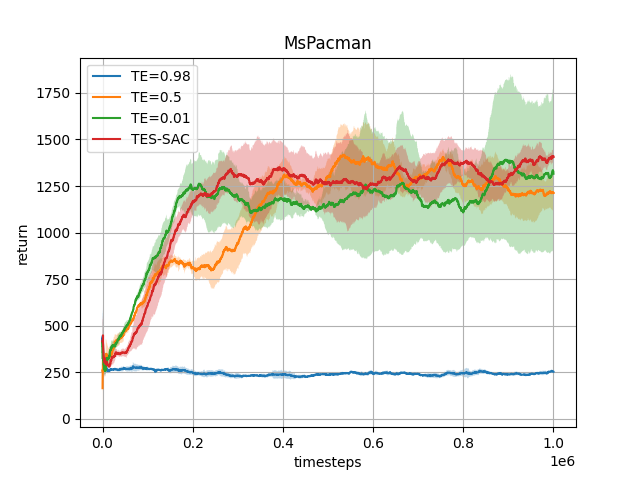}
            \caption{Average return of SAC with different constant $\bar{\mathcal{H}}$ and scheduled $\bar{\mathcal{H}}$ for Atari games BankHeist, Hero, Krull, and MsPacman.} 
            \label{fig:returns}
        \end{figure}

    Figure \ref{fig:averaged_normalized} shows the normalized results over different Atari games, and Figure \ref{fig:returns} shows learning curves for 4 environments. We also include a performance table for all 24 environments we experimented on in Appendix \ref{performanceTable}. We find that TES-SAC significantly outperforms constant target entropy in Hero, BattleZone, BankHeist, and LunarLander. In environments including Assault, MsPacman, Freeway, and Krull, TES-SAC performs similar to the best fixed target entropy SAC. Our Target Entropy Scheduled SAC outperforms constant target entropy in 8 out of 24 environments, which is greater than any other constant target entropy. Theses results suggest that target entropy scheduling can learn better in some environments, and has hyperparameters that are more robust to different domain dynamics than constant target entropy.

    \subsection{Analysis}
    \label{analysis}
        \begin{figure}[H]
                \centering
                \subcaptionbox{\label{sfig:a}}{\includegraphics[width=0.24\textwidth]{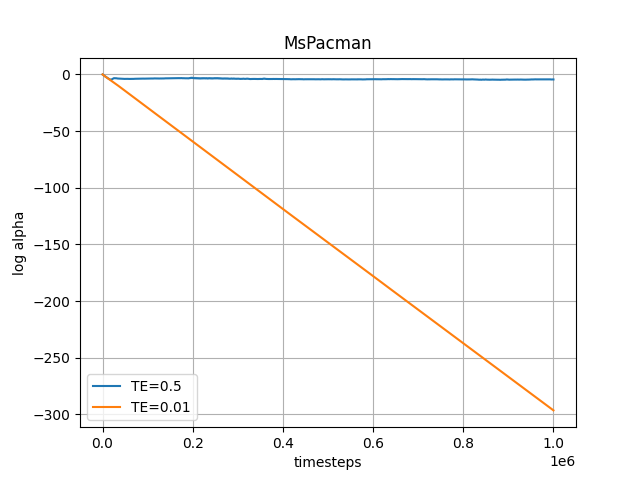}}
                \subcaptionbox{\label{sfig:a}}{\includegraphics[width=0.25\textwidth]{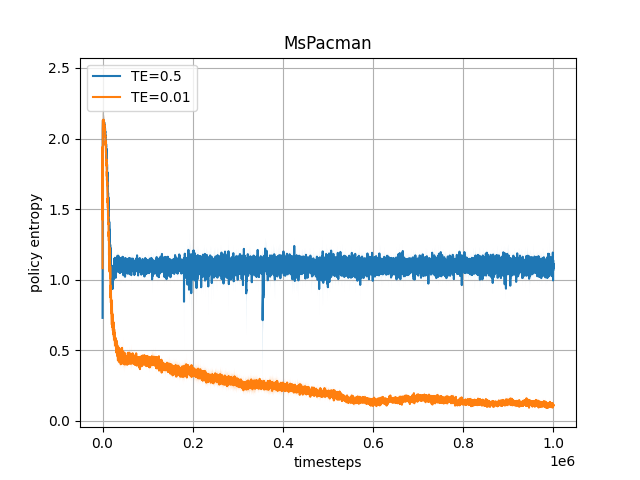}}
                \subcaptionbox{\label{sfig:a}}{\includegraphics[width=0.24\textwidth]{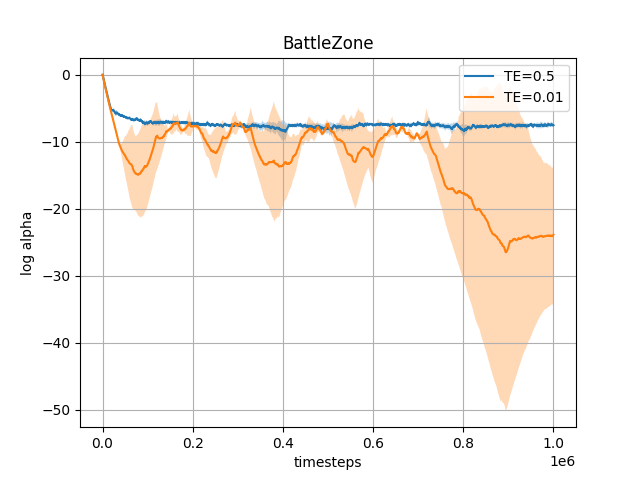}}
                \subcaptionbox{\label{sfig:a}}{\includegraphics[width=0.25\textwidth]{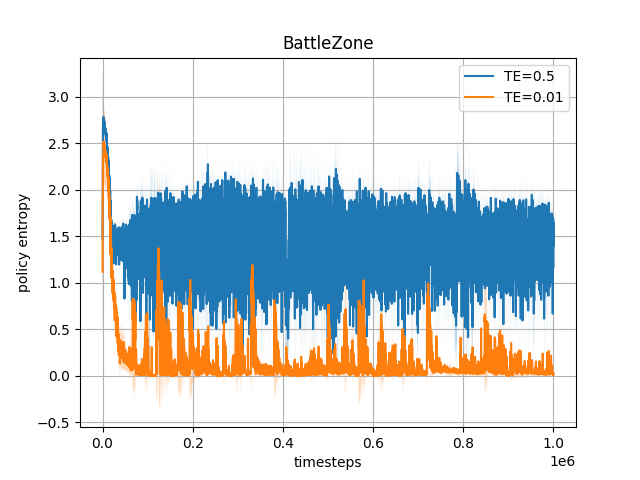}}
                \caption{$\alpha$ dynamics (log scale) and policy entropy under different constant target entropy configurations ($\bar{\mathcal{H}}= 0.01$ and $0.5$) in MsPacman and Battlezone.}
                \label{fig:MsPacman_log_alpha}
        \end{figure}

        The changes in the temperature $\alpha$ determines SAC's performance. Intuitively, a high target entropy induces $\alpha$ to increase and leads to a more stochastic policy, and vice versa. In this section we visualize the $\alpha$ dynamics under different constant target entropy configurations using environments MsPacman and Battlezone. Figure \ref{fig:MsPacman_log_alpha} shows $\log \alpha$ and the policy entropy comparison of low target entropy ($\bar{\mathcal{H}}$ = $0.01 \cdot \log |\mathcal{A}|$) and high target entropy ($\bar{\mathcal{H}}$ = $0.5 \cdot \log |\mathcal{A}|$) for MsPacman and BattleZone. In MsPacman, (1) temperature $\alpha$ exponentially decreases when $\bar{\mathcal{H}} = 0.01 \cdot \log |A|$, and (2) $\alpha$ slowly decreases when $\bar{\mathcal{H}} = 0.5 \cdot \log |A|$. (1) happens because the policy entropy cannot drop to the target entropy. This makes sense in that some environments have a minimum entropy level not close to zero, indicating that the environment has some actions that are really similar --- the impact to the environment from taking one of them is barely different than taking another. For those environments whose minimum entropy is above the target entropy ($\bar{\mathcal{H}}=0.01\cdot \log |\mathcal{A}|$ in our experiments), the policy entropy will never reach the target entropy, which makes the temperature $\alpha$ drop exponentially fast. In MsPacman which has 9 actions, for example, the target entropy is $\bar{\mathcal{H}} = 0.01 \cdot \log 9 = 0.022$. The policy entropy, however, approaches 0.1 without decreasing further, which makes $\alpha$ keep decreasing. This minimum entropy level can vary greatly among different environments. BattleZone is an example the minimum entropy level is close to zero, so even if we set $\bar{\mathcal{H}}$ = $0.01 \cdot \log |\mathcal{A}|$, $\alpha$ does not exponentially decrease like in MsPacman. When the algorithm can quickly reach the minimum entropy level, $\alpha$ starts decreasing rather slowly. This usually happens when we set a slightly larger fixed target entropy, for example $0.5 \cdot \log |\mathcal{A}|$ as used in our experiments. In these cases, policy entropy successfully reaches the target entropy. Without dropping the target entropy, we do not see long-term $\alpha$ decrease. These two $\alpha$ patterns, of course, are sensitive to environment dynamics. Our scheduling tends to heuristically choose the better $\alpha$ pattern for each environment. This makes our scheduling much more robust and domain-generalized.

    \subsection{Ablation Studies}
        \subsubsection{Different Standard Deviation Thresholds}
            \begin{figure}[H]
                \centering
                \subcaptionbox{\label{sfig:a}}{\includegraphics[width=0.4\textwidth]{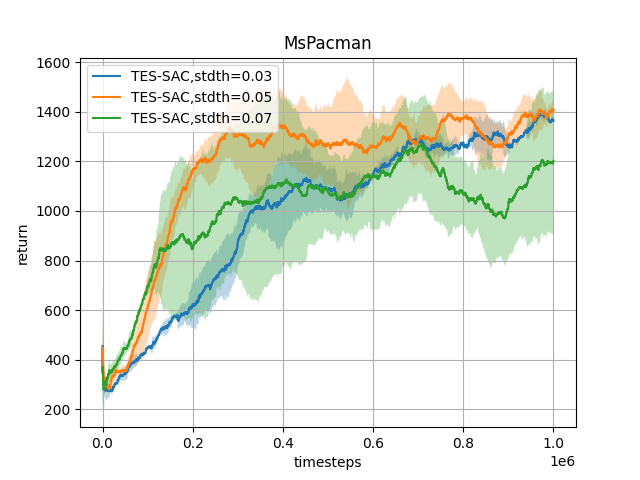}}
                \subcaptionbox{\label{sfig:a}}{\includegraphics[width=0.4\textwidth]{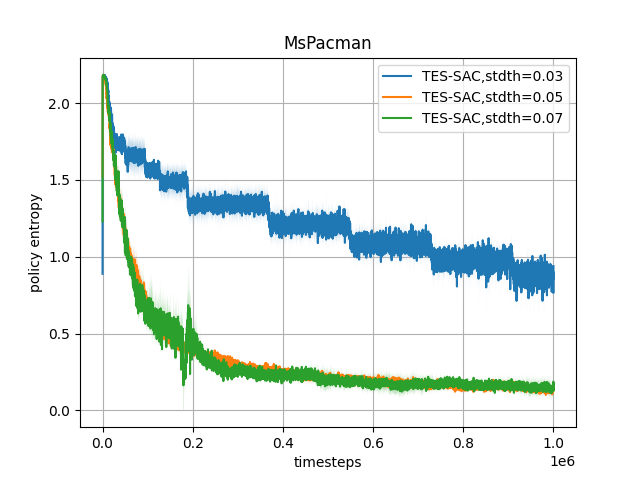}}
                \caption{MsPacman (a) average return and (b) policy entropy, using TES-SAC with std-dev threshold = 0.03, 0.05, 0.07}
                \label{fig:stdth}
            \end{figure}
            In our experiments, we observe that the standard deviation(std) threshold is a relatively more sensitive hyperparameter compared to average threshold and the target decrease factor. The speed at which the target entropy decreases is largely dependant on the standard deviation threshold. We design an experiment to show how different std thresholds will affect the performance. Figure \ref{fig:stdth} is an example in MsPacman, where we use three different std thresholds: $0.03, 0.05,$ and $0.07$. We can see that different std thresholds do change the speed of target entropy decrease, but don't actually affect the performance much. This suggests that our hyperparameters require little tuning.
            
        \subsubsection{Fixed-step Scheduling V.S. Exponential Window Scheduling}
            \begin{figure}[H]
                \centering
                \subcaptionbox{\label{sfig:a}}{\includegraphics[width=0.31\textwidth]{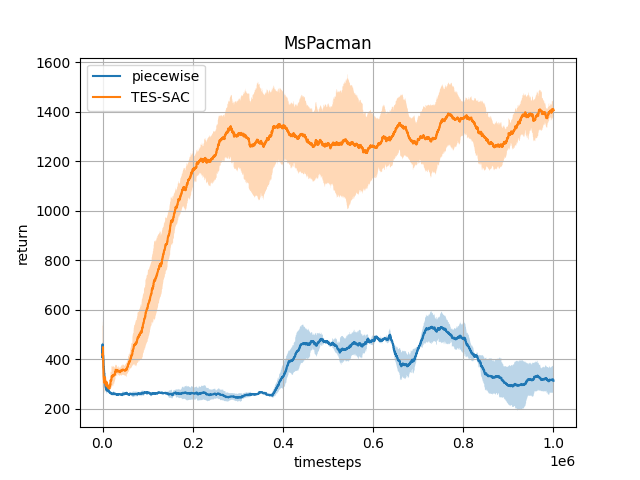}}
                \subcaptionbox{\label{sfig:a}}{\includegraphics[width=0.31\textwidth]{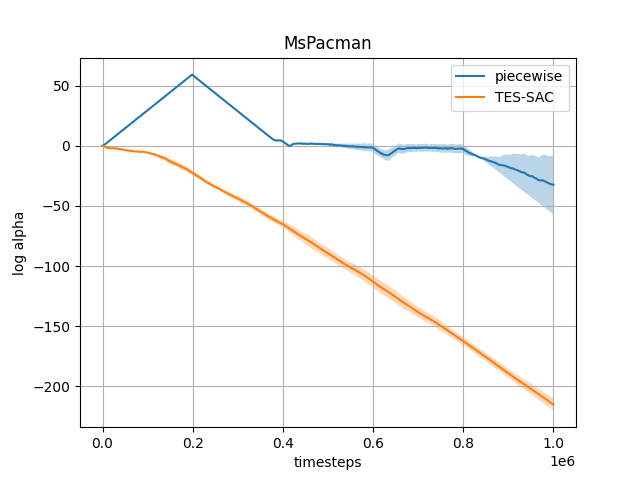}}
                \subcaptionbox{\label{sfig:a}}{\includegraphics[width=0.31\textwidth]{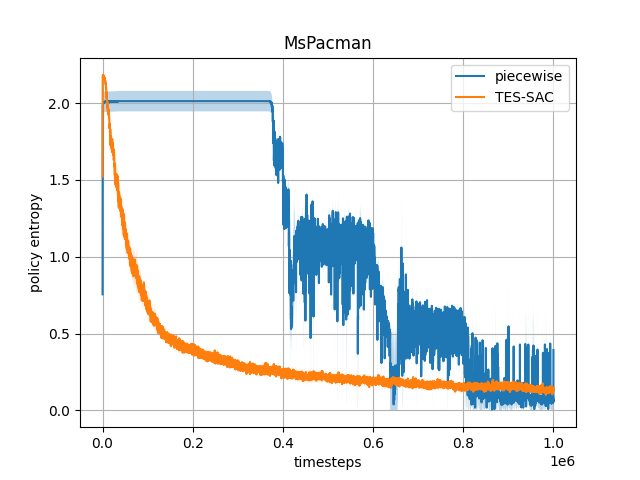}}
                \caption{MsPacman Fixed-step V.S. Exp Schedule, (a) average return, (b) $\log \alpha$, (c) policy entropy} 
                \label{fig:piecewise}
            \end{figure}
           We compare fixed-step scheduling with our scheduling method to justify why we use policy entropy as a signal for dropping the target entropy. Fixed-step scheduling drops the target entropy every certain number of experience steps. This means fixed-step scheduling requires potentially heavy tuning for best performance. Figure \ref{fig:piecewise} shows the average return, $\log \alpha$ value, and policy entropy in the MsPacman environment when we evenly drop the target entropy four times within 1M experience steps to be $[0.98\log|\mathcal{A}|, 0.75\log|\mathcal{A}|, 0.5\log|\mathcal{A}|, 0.25\log|\mathcal{A}|, 0.01\log|\mathcal{A}|]$.  TES-SAC significantly outperforms fixed-step scheduling, and the $\alpha$ value under fixed-step scheduling becomes ridiculously high in early training.
            
            We share our ablation study results on more environments in Appendix C.
            
    \subsection{Applying Temperature Scheduling to Soft Q-Learning}
    \label{discussions}
    The scheduling method we proposed for SAC can potentially be applied to Soft Q-Learning as well. In SQL, we approximate the policy as a Q-Function-based soft-greedy policy \citep{Fox2016Taming}:
    \begin{equation}
        \label{softmaxp}
        \pi(a|s) = \frac{\exp(\frac{1}{\alpha}Q(s, a))}{\sum_a{\exp(\frac{1}{\alpha}Q(s, a))}}
    \end{equation}
    
    We can then apply the target entropy scheduling in Section \ref{expSchedule} to calculate $\alpha$ using the temperature Lagrange term with this soft-greedy policy similar to \eqref{alphaloss}:
    \begin{equation}
        J(\alpha) = \E_{a\sim \pi}\Bigg[-Q(s, a) + \alpha \log \sum_{a'}\exp \frac{1}{\alpha}Q(s, a') - \alpha \bar{\mathcal{H}}\Bigg].
    \end{equation}
    
\begin{figure}[H]
        \label{f1}
            \centering
            \subcaptionbox{\label{sfig:a}}{\includegraphics[width=0.24\textwidth]{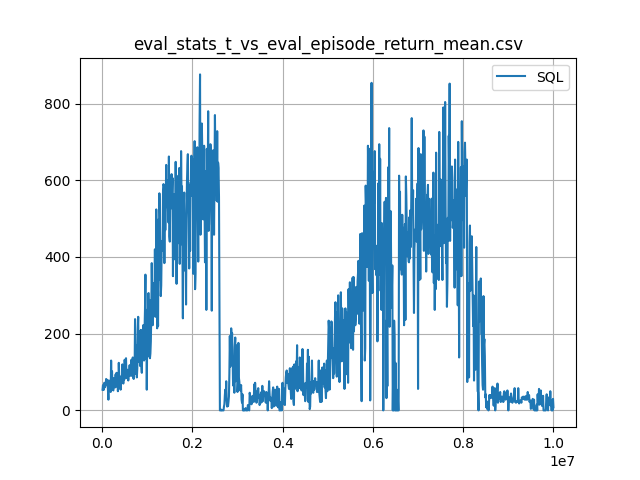}}
            \subcaptionbox{\label{sfig:a}}{\includegraphics[width=0.24\textwidth]{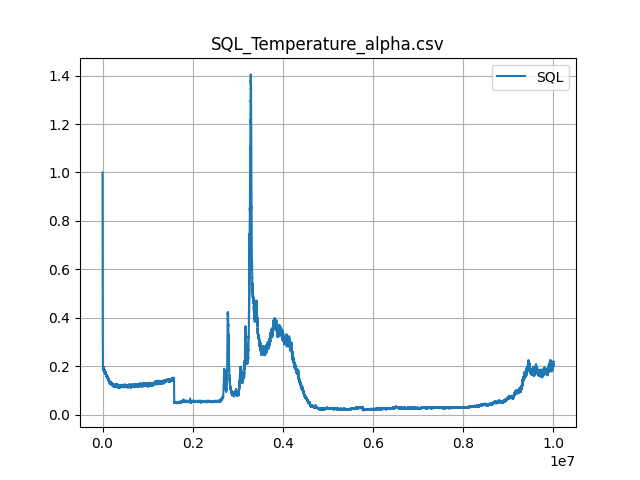}}
            \subcaptionbox{\label{sfig:a}}{\includegraphics[width=0.24\textwidth]{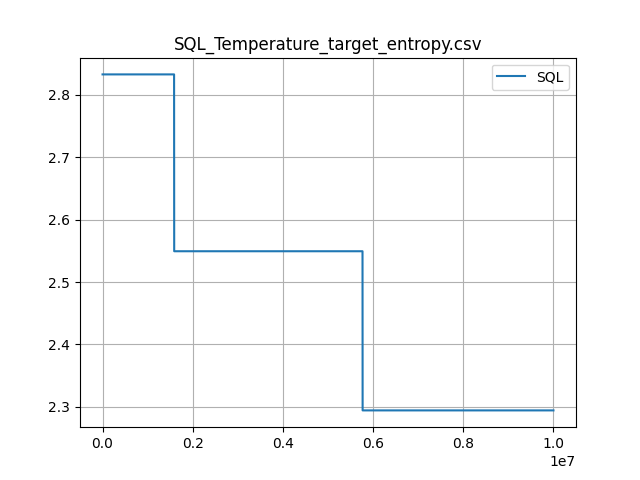}}
            \subcaptionbox{\label{sfig:a}}{\includegraphics[width=0.24\textwidth]{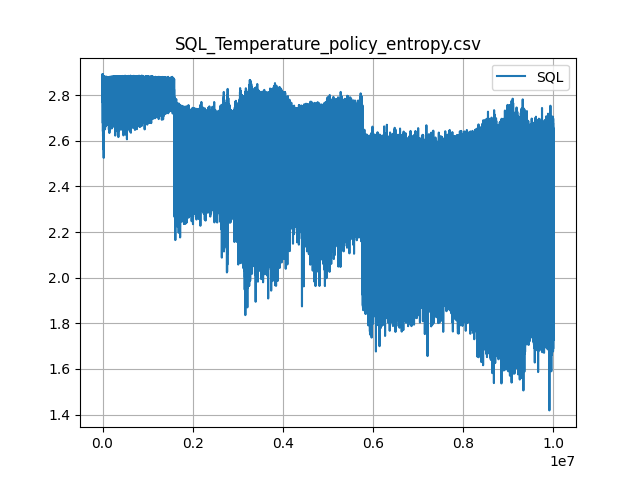}}
            \caption{(a) episode return mean, (b) temperature $\alpha$, (c) target entropy, and (d) policy entropy of Seaquest, using SQL with target entropy scheduling. The reward drops to zero in (a) after dropping the target entropy, visulaizing the instability.} 
            \label{fig:sql}
        \end{figure}

    A Q-Function-based policy is very unstable when adjusting the target entropy, however, because $\alpha$ will abruptly decrease when the target entropy drops. This will cause a sudden shift in the soft-greedy policy described in \eqref{softmaxp}. Figure \ref{fig:sql} visualizes this instability using the Atari 2600 game Seaquest as an example. 
    
        \label{section:SAC_exp}
    The cause for this difference in applying our target entropy scheduling to SAC and SQL is that in SAC there is a policy network $\pi_{\phi}$ instead of a Q-Function-based soft-greedy policy. This means we can directly calculate the policy entropy using \eqref{entropy} with the policy network $\pi_{\phi}(a|s)$. A policy network will effectively stabilize the system, and somewhat tolerates the fast change in temperature since it will take some time to learn this change.

\section{Conclusion}
    In this article, we analyze some issues with applying SAC to discrete action spaces with a constant target entropy: such a target entropy would be extremely environment dependent and require fine-tuning. We present TES-SAC as an alternative, a heuristic method to schedule the target entropy for discrete Soft Actor Critic by observing the dynamics in the policy entropy. We show empirical evidence that this scheduling method requires little tuning, is more robust, and generally outperforms SAC with a constant target entropy. We also explain why this target entropy scheduling method will not be as effective when applied to SQL. We have not yet attempted to tune our method towards state-of-the-art performance, because our scheduling is rather heuristic and is not backed by strong theory to make every part incontrovertible. Nevertheless, we believe this work is an important step forward in terms of annealing the temperature in soft actor critic.

    For future work, we are interested in applying a similar method to continuous SAC target entropy scheduling. Although continuous SAC already demonstrates state-of-the-art performance using a constant target entropy, our results suggest that integrating target entropy scheduling could potentially improve the performance.


\bibliography{main}
\bibliographystyle{iclr2022_conference}

\newpage
\appendix
\section{Hyperparameters}
    \begin{table}[H]
    \caption{Hyperparameter for Discrete SAC with Exponential Window Scheduling}
    \label{hyperparameters}
    \begin{center}
    \begin{tabular}{ll}
    \multicolumn{1}{c}{\bf Hyperparameter} 
        &\multicolumn{1}{c}{\bf Value}
    \\ \hline \\  
    learning rate & $3\times10^{-4}$ \\ 
    optimizer & Adam \\
    mini-batch size & 256 \\
    discount ($\gamma$) & 0.99 \\
    buffer size & $10^5$ \\
    hidden layers & 2 \\
    hidden units per layer & 512 \\
    activation function & ReLU \\
    target smoothing coefficient ($\tau$) & 0.005 \\
    target update interval & 1 \\
    gradient steps & 1 \\
    average threshold & 0.01 \\
    standard deviation threshold & 0.05 \\
    target entropy discount & 0.9 \\
    exponential window discount $\lambda$ & 0.999 \\
    \end{tabular}
    \end{center}
    \end{table}
    
\section{Algorithm for Target Entropy Schedule}
\label{scheduling}
    \begin{algorithm}[H]
    \caption{Target Entropy Schedule}
    \label{alg:algorithm}
    \textbf{Input}: current policy entropy: $e_t$\\
    \textbf{Parameters}: 
                        exponential window discount $\lambda$, 
                        avg\_threshold $\bar{\mu}$, 
                        std\_threshold $\bar{\sigma}$, \\
                        discount factor $k$, 
                        total conditioned num $T$, 
                        initial target entropy $\bar{\mathcal{H}}_0$.\\
    \textbf{Output}: current target entropy $\bar{\mathcal{H}}$
    \begin{algorithmic}[1] 
    \STATE Let $\hat{\mu}$ = $\bar{\mathcal{H}}_0$, $\hat{\sigma}$ = 0, $i$ = 0, $\bar{\mathcal{H}} = \bar{\mathcal{H}}_0$
    \FOR{each timestep}
    \STATE $\delta = e_t - \hat{\mu}$
    \STATE $\hat{\mu} = \hat{\mu} + (1-\lambda) * \delta$
    \STATE $\hat{\sigma}^2 = \lambda * (\hat{\sigma}^2 + (1-\lambda) * \delta^2)$
    \STATE $\hat{\sigma} = \sqrt{\hat{\sigma}^2}$
    \IF {not ($\bar{\mathcal{H}} - \bar{\mu}$ $<$ $\hat{\mu}$ $<$ $\bar{\mathcal{H}} + \bar{\mu}$) or $\hat{\sigma}$ $>$ $\bar{\sigma}$}
    \STATE \textbf{return} $\bar{\mathcal{H}}$
    \ENDIF
    \STATE $i = i + 1$
    \IF {$i$ $\geq$ $T$}
    \STATE $i = 0$
    \STATE $\bar{\mathcal{H}} = \bar{\mathcal{H}} * k$
    \STATE \textbf{return} $\bar{\mathcal{H}}$
    \ENDIF
    \ENDFOR
    \end{algorithmic}
    \end{algorithm}

\section{Ablation Study on more environments}
    \subsection{Different Standard Deviation Thresholds}
        \begin{figure}[H]
                \centering
                \includegraphics[width=0.4\textwidth]{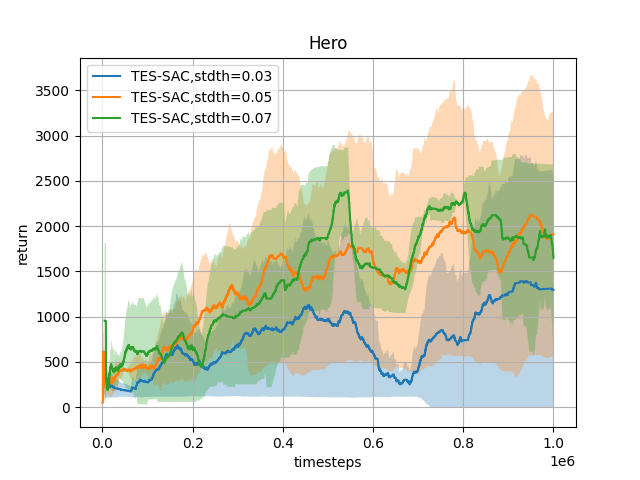}
                \includegraphics[width=0.4\textwidth]{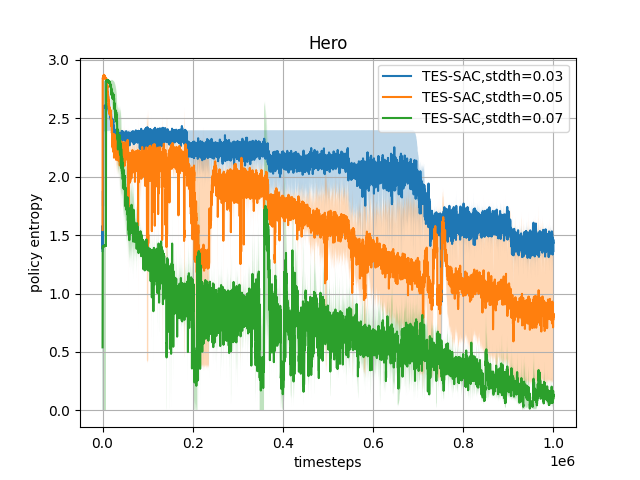}
                \subcaptionbox{\label{sfig:a}}{\includegraphics[width=0.4\textwidth]{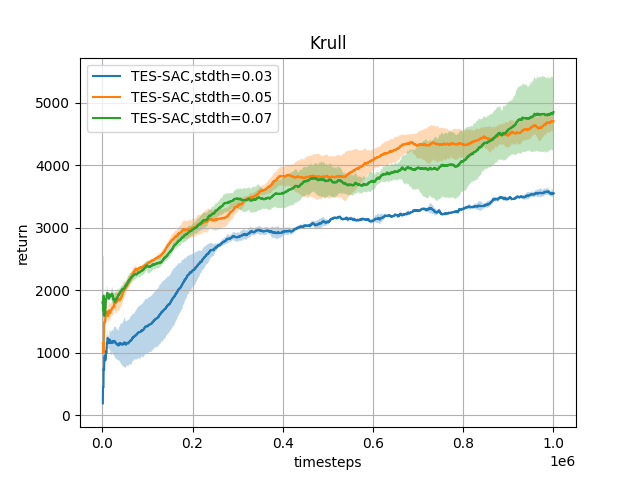}}
                \subcaptionbox{\label{sfig:a}}{\includegraphics[width=0.4\textwidth]{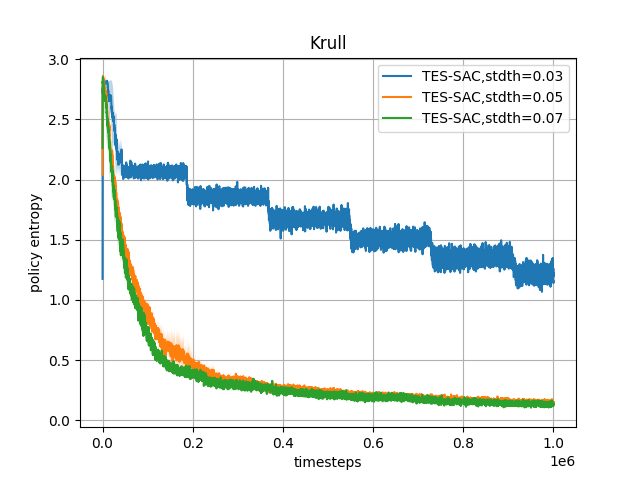}}
                \caption{TES-SAC with std threshold = 0.03, 0.05, 0.07 on Hero and Krull, (a) is average return , and (b) is policy entropy}
                \label{fig:my_label}
        \end{figure}
    
    \subsection{Fixed-step Scheduling V.S. Exponential Window Scheduling}
        \begin{figure}[H]
                \centering
                \includegraphics[width=0.32\textwidth]{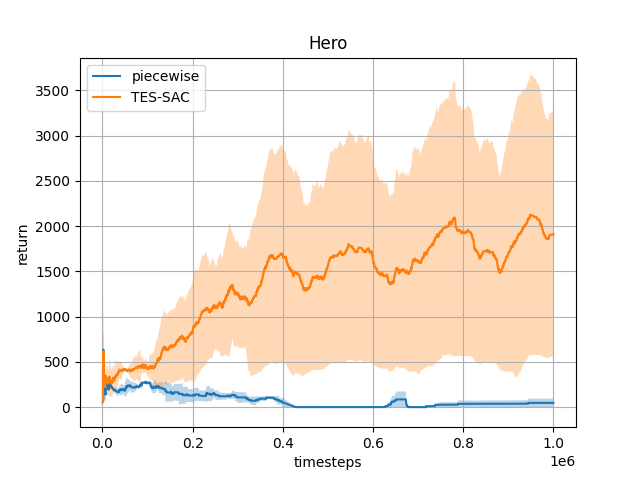}
                \includegraphics[width=0.32\textwidth]{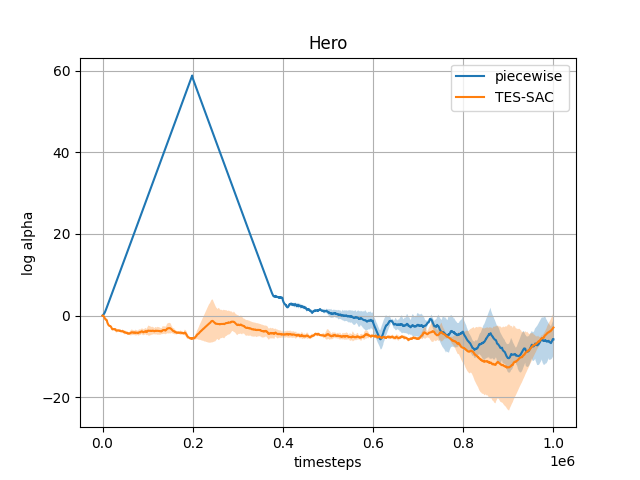}
                \includegraphics[width=0.32\textwidth]{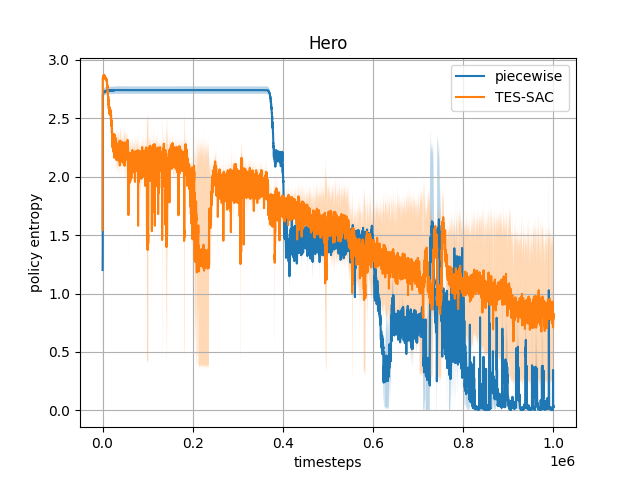}
                \subcaptionbox{\label{sfig:a}}{\includegraphics[width=0.32\textwidth]{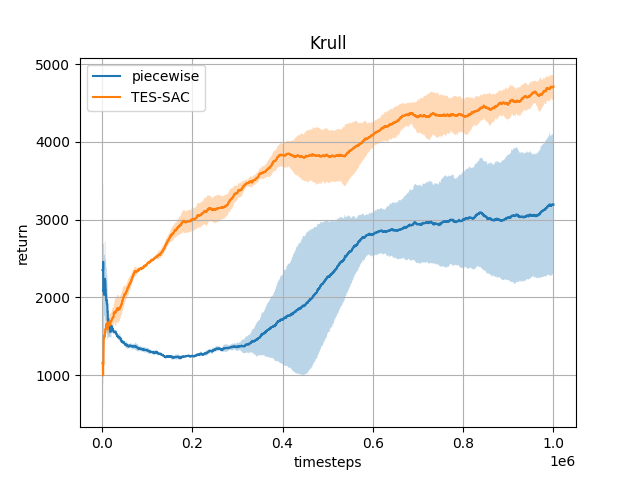}}
                \subcaptionbox{\label{sfig:a}}{\includegraphics[width=0.32\textwidth]{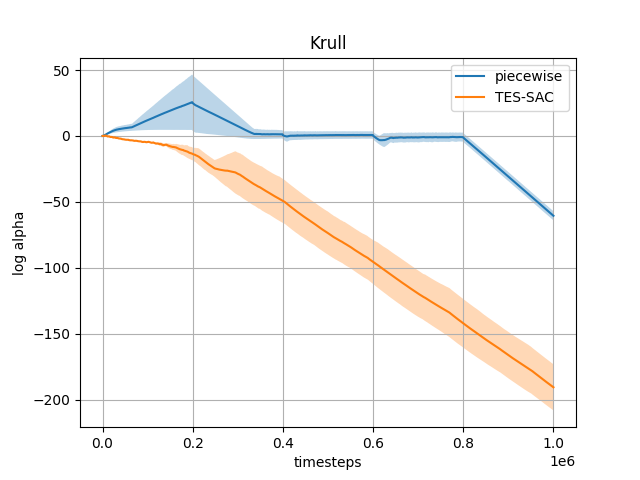}}
                \subcaptionbox{\label{sfig:a}}{\includegraphics[width=0.32\textwidth]{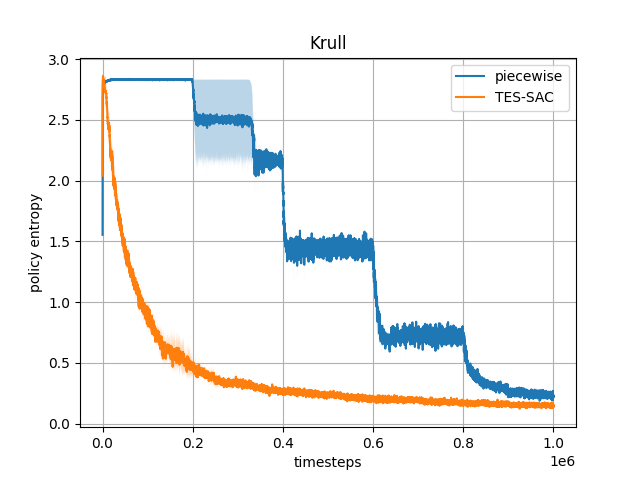}}
                \caption{Piecewise V.S. Exp Schedule on Hero and Krull, (a) is average return, (b) is $\log \alpha$, and (c) is policy entropy} 
        \end{figure}
    
\section{Normalized performance}
    \begin{figure}[H]
        \centering
        \includegraphics[width=0.9\textwidth]{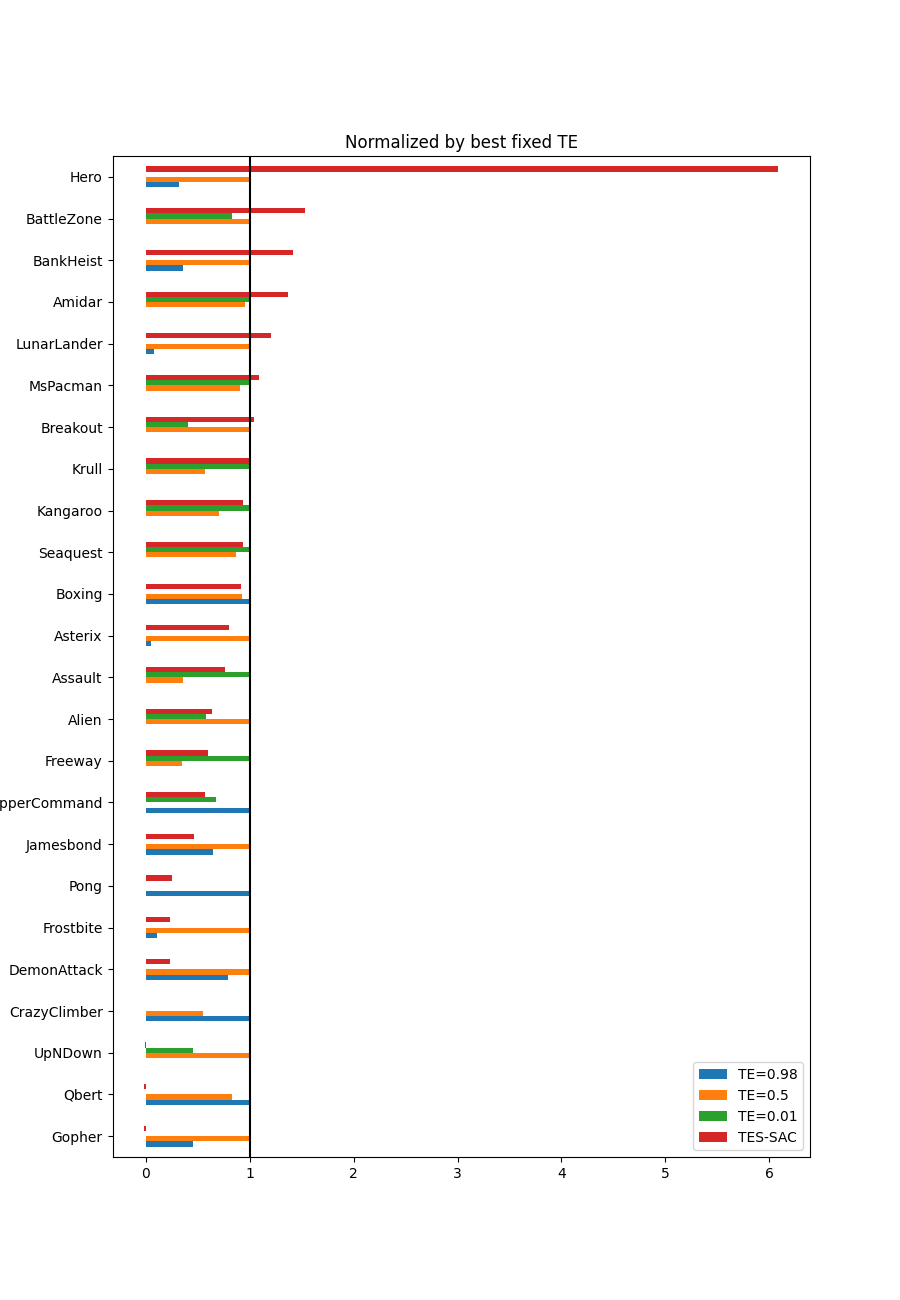}
        \caption{Performance of TES-SAC normalized by best-performed fixed constant target entropy SAC, using the formula (algorithm@1M - worstfixedTE\_SAC\_score) / (bestfixedTE\_SAC\_score - worstfixedTE\_SAC\_score)}
        \label{fig:normalized}
    \end{figure}

\newpage

\section{Performance Table}
\label{performanceTable}

    \begin{table}[H]
        \label{sample-table}

        \begin{center}
        \scalebox{0.85}{
        \begin{tabular}{lllllllll}
        \multicolumn{1}{c}{Environment}  
            &\multicolumn{1}{l}{C=0.98} 
            &\multicolumn{1}{l}{}
            &\multicolumn{1}{l}{C=0.5} 
            &\multicolumn{1}{l}{}
            &\multicolumn{1}{l}{C=0.01}
            &\multicolumn{1}{l}{}
            &\multicolumn{1}{l}{TES-SAC}
            &\multicolumn{1}{l}{}
        \\ \hline \\  
         Lunarlander         & -217.63 & (8.21)                      & 187.17 &(57.81)                           & -251.57 &(623.25)                  & \textbf{277.43} &(5.46)     \\
         Alien               & 196.70 &(11.43)                       & \textbf{996.07} & (383.50)                & 640.77 & (454.88)                  & 685.93 & (101.94)           \\         
         Amidar              & 1.36 &(0.47)                          & 29.65 &(17.19)                            & 31.14 &(11.70)                     & \textbf{42.07} &(26.97)     \\  
         Assault             & 245.07 &(5.14)                        & 288.97 &(108.25)                          & \textbf{366.43} &(13.56)           & 337.03 &(19.15)             \\          
         Asterix             & 220.67 &(17.46)                       & \textbf{421.83} &(93.77)                  & 260.83 &(187.53)                   & 378.5 &(75.31)              \\      
         BankHeist           & 10.57 &(1.86)                         & 25.63 &(15.25)                            & 4.57 &(4.98)                       & \textbf{35.40} &(8.88)      \\      
         BattleZone          & 3816.67 &(106.25)                     & 5103.33 &(1649.05)                        & 4663.33 &(359.29)                  & \textbf{5790.0} &(2016.80)  \\
         Boxing              & \textbf{0.64} &(0.23)                 & -1.58 &(0.99)                             & -28.5 &(9.86)                      & -1.87 &(2.97)               \\         
         Breakout            & 1.32 &(0.18)                          & 2.60 &(1.50)                              & 2.37 &(0.75)                       & \textbf{2.65} &(0.02)       \\       
         ChopperCommand      & \textbf{802.17} &(112.06)             & 206.67 &(45.84)                           & 590.5 &(290.63)                    & 532.73 &(310.68)            \\          
         CrazyClimber        & \textbf{3560.67} &(90.48)             & 2036.33 &(916.45)                         & 113.0 &(80.45)                     & 4.0 &(4.97)                 \\          
         DemonAttack         & 165.4 &(4.80)                         & \textbf{182.93} &(69.62)                  & 99.55 &(10.76)                     & 157.20 &(2.90)              \\          
         Freeway             & 0.82 &(0.03)                          & 8.26 &(6.59)                              & \textbf{14.63} &(10.35)            & 13.57 &(3.85)               \\          
         Frostbite           & 54.43 &(13.48)                        & \textbf{245.47} &(65.15)                  & 32.03 &(24.84)                     & 81.03 &(91.64)              \\         
         Gopher              & 273.0 &(141.97)                       & \textbf{344.40} &(37.42)                  & 210.80 &(62.51)                    & 154.87 &(142.72)            \\         
         Hero                & 289.77 &(194.12)                      & 481.25 &(481.39)                          & 200.60 &(299.72)                   & \textbf{1908.67} &(1352.89) \\     
         Jamesbond           & 41.67 &(6.33)                         & \textbf{60.67} &(25.93)                   & 6.33 &(8.96)                       & 31.33 &(13.82)              \\          
         Kangaroo            & 45.33 &(8.05)                         & 241.33 &(164.09)                          & \textbf{325.33} &(241.01)          & 307.33 &(162.50)            \\       
         Krull               & 1647.0 &(54.79)                       & 3383.67 &(59.67)                          & 4716.66 &(557.11)                  & \textbf{4717.67} &(150.67)  \\
         MsPacman            & 253.20 &(1.98)                        & 1213.33 &(97.77)                          & 1315.83 &(410.08)                  & \textbf{1408.0} &(39.10)    \\ 
         Pong                & \textbf{-20.36} &(0.37)               & -21 &(0)                                  & -21 &(0)                           & -20.84 &(0.21)              \\          
         Qbert               & \textbf{222.03} &(31.93)              & 203.75 &(252.98)                          & 115.10 &(94.22)                    & 74.93 &(28.29)              \\         
         Seaquest            & 37.53 &(7.26)                         & 169.73 &(28.31)                           & \textbf{69.80} &(61.89)            & 116.73 & (72.77)            \\    
         UpNDown             & 325.23 &(134.03)                      & \textbf{1185.77} &(699.92)                & 715.93 &(885.67)                   & 207.6 &(242.48)             \\        
        \end{tabular}
        }
        \end{center}
        \caption{Performance at 1M interactions. The results of constant $\bar{\mathcal{H}}$ SAC and TES-SAC show the average score over three runs, with standard deviation in the parenthesis. First three columns are constant target entropy $\bar{\mathcal{H}}=C * log|A|$ with $C = 0.98$, $C = 0.5$, and $C = 0.01$. The forth columns is our TES-SAC. \citep{DBLP:journals/corr/abs-1712-09381}. Best performance is shown in bold.}
    \end{table}

\end{document}